\def\year{2020}\relax
\documentclass[letterpaper]{article} 
\usepackage{aaai20}  
\usepackage{times}  
\usepackage{helvet} 
\usepackage{courier}  
\usepackage[hyphens]{url}  
\usepackage{graphicx} 
\usepackage{amsmath}
\usepackage{mathtools} 
\usepackage{graphicx}  
\title{Facial Expression Recognition Using Disentangled Adversarial Learning }
\author{Kamran Ali, Charles E. Hughes\\
Synthetic Reality Lab, Department of Computer Science\\
University of Central Florida Orlando, Florida\\
kamran@knights.ucf.edu, ceh@cs.ucf.edu 
}

\begin{document}

\maketitle

\begin{abstract}
The representation used for Facial Expression Recognition (FER) usually contain expression information along with other variations such as identity and illumination. In this paper, we propose a novel Disentangled Expression learning-Generative Adversarial Network (DE-GAN) to explicitly disentangle facial expression representation from identity information. In this learning by reconstruction method, facial expression representation is learned by reconstructing an expression image employing an encoder-decoder based generator. This expression representation is disentangled from identity component by explicitly providing the identity code to the decoder part of DE-GAN. The process of expression image reconstruction and disentangled expression representation learning is improved by performing expression and identity classification in the discriminator of DE-GAN. The disentangled facial expression representation is then used for facial expression recognition employing simple classifiers like SVM or MLP. The experiments are performed on publicly available and widely used face expression databases (CK+, MMI, Oulu-CASIA). The experimental results show that the proposed technique produces comparable results with state-of-the-art methods.  
\end{abstract}

\section{Introduction}
Facial expression recognition (FER) has many exciting applications in domains like human-machine interaction,  intelligent tutoring system (ITS), analysis and diagnosis of suicidal symptoms, interactive games, and intelligent transportation. Therefore FER has been widely studied by computer vision and machine learning community in the past decades.  Despite all the research, FER is still a difficult and challenging task for the research community. Most of FER techniques developed so far, do not consider inter-subject variations and differences in facial attributes of individuals present in data. An explicit disentanglement of facial expression features from identity representation has been studied and explored in very few FER techniques \cite{r4} \cite{r5}. In most of FER techniques in the literature, the representations used for the classification of expressions contain identity-related information along with facial expression information, as observed in \cite{r1} \cite{r2}. The main drawback of this entangled representation is that it negatively affects the generalization capability of FER technique, which, as a result, degrades the performance of FER on unseen identities.

\begin{figure}[t]
    \centering
    \includegraphics[width=8cm,, height=5.5cm]{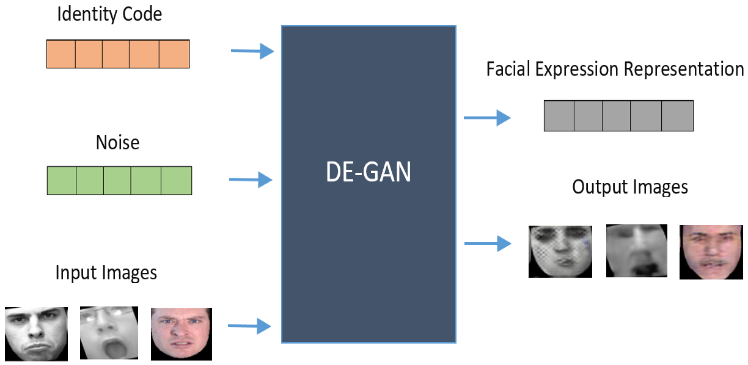}
    \caption{DE-GAN takes input image(s), random noise vector and identity code as input, and outputs a disentangled facial expression representation along with synthetic image(s) with the same facial expression(s) as in the input image(s), but with different identity, specified by the identity code. The disentangled facial expression representation is then used for FER.}
    \label{fig:1}
\end{figure}

In this paper, we present a GAN based expression representation learning technique which is inspired by the disentangled adversarial loss presented in \cite{r34}. The proposed Disentangled Expression-learning Generative Adversarial Network, which we call DE-GAN, is used to learn an expression representation which does not contain identity information. The main objective of DE-GAN is to extract expression information  $ f_{exp} (x) $ from non-linear function, $h$, defined as: $f(x) = h(f_{id} (x), f_{exp} (x))$, where $f(x)$ is the representation of an image, $f_{exp} (x)$ corresponds to facial expression representation, and $f_{id} (x)$ denotes the identity specific representation. The architecture of our proposed DE-GAN is shown in Figure \ref{fig:1}.

The motivation to learn a discriminative representation for FER encourages us to employ an encoder-decoder based GAN model. An image containing any basic expression (i.e anger, disgust, contempt, fear, happy, sad and surprise) is fed into the encoder, $G_{en}$, and a reconstructed image containing the same facial expression as the input image is generated from the decoder $G_{de}$ with a target identity. So, generator, $G$, is used for two purposes: 1. to learn a generative and discriminative facial expression representation, which is the output of the encoder, $G_{en}$, 2. to reconstruct a facial expression image, which is the output of the decoder $G_{de}$. The discriminator, $D$, of DE-GAN is trained to classify not only between real and fake images but to also perform the classification of identities and facial expressions. The estimation of facial expressions and identities in the discriminator helps us in constructing a generator which is capable of generating a discriminative and disentangled facial expression representation from input facial expression images. The multi-task classification in the discriminator also plays an important role in the reconstruction of facial expression images using the expression representation encoded by the encoder $G_{en}$ and the target identity provided to the decoder $G_{de}$ of DE-GAN.  

Our generator, $G$, and discriminator, $D$, are quite different from $G$ and $D$ used in conventional GAN \cite{r3}. In \cite{r3}, the input to the generator is random noise and the output is a synthetic image. While in our case, $G$ is fed with a facial expression image $x$, an identity code $I$ and random noise $z$, and the output of $G$ is a reconstructed image with the same facial expression as in $x$, but with a different identity, specified by an identity code, $I$. This synthesized image is then used to fool the discriminator, $D$. The noise vector $z$ and identity code $I$ is concatenated with the facial expression representation learned by the encoder $G_{en}$. This concatenated vector is then fed into the decoder $G_{de}$ to reconstruct a facial expression image. The disentangled facial expression representation learned by the encoder $G_{en}$, is mutually exclusive from identity information, which can be best used for FER.

\vspace{2mm}
In contrast to previous methods \cite{r4} \cite{r5} which employ an expression-sensitive contrastive loss, an identity-sensitive contrastive loss and learning de-expression residue from generative model, to reduce the influence of identity related information from expression representations, our proposed DE-GAN based technique learns an explicitly disentangled facial expression representation which can be used to improve FER.

\vspace{2mm}
The main contributions of this paper are as follows:

\begin{itemize}
\item We present a novel disentangled and discriminative facial expression representation learning technique for FER using adversial learning in DE-GAN framework.
\item The DE-GAN based set-up also performs the task of facial expression synthesis by transferring facial expression information from input image to target identity.
\end{itemize}

\section{Related Work}
Facial expression recognition has been extensively studied in the past decades, as elaborated in recent surveys \cite{r6} \cite{r7} \cite{r8}. The main goal of FER is to extract features that are discriminative and invariant to variations such as pose, illumination, and identity-related information. This feature extraction process can be divided into two main categories: human-engineered features and learned features. Before deep learning era, most of FER techniques involved human-designed features using techniques such as Histograms of Oriented Gradients (HOG) \cite{r9}, Scale Invariant Feature Transform (SIFT) features \cite{r10}, \cite{r11}, histograms of Local Binary Patterns (LBP) \cite{r12}, \cite{r13}, histograms of Local Phase Quantization (LPQ) \cite{r14}. These techniques are applied to static images or a sequence of images to extract features for FER.

\vspace{2mm}

The human-crafted features perform well in lab controlled environment where the expressions are posed by the subjects with constant illumination and stable head pose. However, these features fail on spontaneous data with varying head position and illumination. Recently, deep CNN \cite{r15}, \cite{r16}, \cite{r17} \cite{r28} \cite{r30} \cite{r32} \cite{r33} have been employed to increase the robustness of FER to real-world scenarios. However, the learned deep representations used for FER are often influenced by large variations in individual facial attributes such as ethnicity, gender, age, etc of subjects involved in training. The main drawback of this phenomenon is that it negatively affects the generalization capability of the model, and as a result, the FER accuracy is degraded on unseen subjects. Although significant progress has been made in improving the performance of FER, the challenge of mitigating the influence of inter-subject variations on FER is still an open area of research. 

\vspace{2mm}

Various techniques \cite{r18}\cite{r19} have been proposed in the literature to increase the discriminative property of extracted features for FER by increasing the inter-class differences and reducing intra-class variations. Most recently, Identity-Aware CNN (IACNN) \cite{r4} was proposed to enhance FER performance by reducing the effect of identity related information by using an expression-sensitive contrastive loss and an identity-sensitive contrastive loss. However, the effectiveness of contrastive loss is affected by large data expansion, which is caused due to the compilation of training data in the form of image pairs \cite{r2}. An Identity-free conditional Generative Adversarial Network (IF-GAN) \cite{r2} method was proposed to mitigate the effect of identity-related information by generating a common synthetic image with the same facial expression as the input image. This generated synthetic image is then used for FER to eliminate the influence of individual variations of subjects in data. The problem with this technique is that since the classification of facial expression is performed on the generated synthetic image, therefore the performance of FER will not only depend on the quality of the generated image but also on the facial expression transfer process from the input image to the synthetic image. In \cite{r5}, person-independent expression representations are learned by using De-expression Residue Learning (DeRL). In DeRL, a cGAN framework is trained to generate a neutral image from an expression image fed to the generator. The expression representation is then learned from inter-mediate layers of the generator, after training the cGAN framework. However, the DeRL based technique, apart from being computationally very costly, does not explicitly disentangle the expression information from identity information, because the same intermediate representation is used to generate neutral images of the same identities. 

\section{Proposed Method}
The proposed facial expression expression recognition technique contains two learning processes, i.e in the first process DE-GAN is employed to learn a disentangled and discriminative expression representation by synthesising an expression image, and the second part involves the facial expression recognition by classifying the disentangled expression features using a simple shallow neural network. The image pairs, e.g $< f_{input}, f_{target}>$ are used to train the DE-GAN. $f_{input}$ is an expression image of any identity, and $f_{target}$ is a synthesised image having the same expression as input image but with different identity. After training, the generator reconstructs an expression image $f_{target}$ by extracting the expression information from $f_{input}$ and transferring it to $f_{target}$. During this learning by reconstruction process, the network learns to disentangle expression features from identity information. In the second phase of learning FER is performed employing a shallow network using the disentangled expression features extracted from the encoder of DE-GAN$'$s generator.

\subsection{Expression Transfer Using DE-GAN}
The generator in DE-GAN is based on encoder-decoder structure, while the discriminator is a simple deep convolutional neural network. The input to the generator is an expression image, which is passed through the encoder part of generator and a disentangled representation vector connects the encoder with the decoder of the generator. This disentangled representation is then used to generate an image with the same expression but with different identity.  

In order to generate an image containing the expression of the input image but with different identity, the expression information must be captured in such a way that it does not contain the identity features of the input image. Because the identity information is fed in the form of Id code to the decoder, which is then combined with the expression information to generate an expression image but with different identity.  Thus, by providing the identity information explicitly to the decoder we will be able to disentangle the expression information from the identity features in the expression representation. 

\subsection{Disentangled Expression Representation}
\begin{figure*}[ht!]
\centering
\includegraphics[width=15cm,, height=5cm]{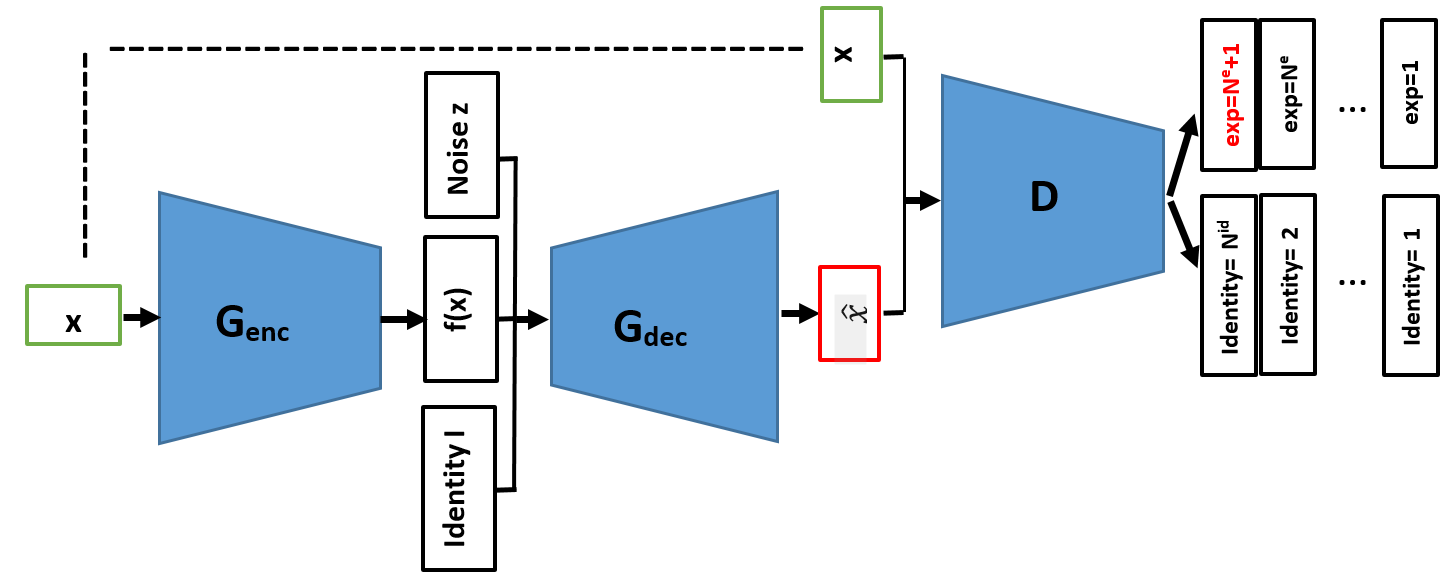}
\caption{Architecture of our DE-GAN}
\label{fig:2}
\end{figure*}

Given a face expression image $x$ with expression label as $y^e$ and identity label as $y^{id}$, our main objective is to learn a discriminative expression representation for FER by generating an expression image $\bar{x}$ with the same expression label as $y^e$ but with a different identity label, lets say $y^{id2}$ by employing DE-GAN. The DE-GAN is an encoder-decoder based conditional GAN which is conditioned on the original image $x$ and the identity label $y^{id}$. The architecture of DE-GAN is shown in Figure \ref{fig:2}.

\subsubsection{Generator: }
Given an expression image $x$, the extracted features $f(x)$ is given by a non-linear function $h$ as:

\begin{align}
f(x) = h(f_{id} (x), f_{exp} (x))
\end{align}

Where $f_{id}$ corresponds to the identity information and $f_{exp} (x)$ denotes the expression features. Our hypothesis is that the accuracy of FER can be improved by disentangling $f_{id} (x)$ from $f_{exp} (x)$. In DE-GAN this disentangling is performed by employing an encoder $G_{en}$ and a decoder $G_{de}$ based generator. The goal of encoder is to learn a disentangled expression representation i.e $f_{exp} (x)= G_{en} (x)$ given a face image x, while decoder $G_{de}$ is used to generate a synthetic expression image given by $\bar{x} = G_{de} (f_{id} (x),I,z)$, where variances like illumination, age, gender etc are modeled by noise $z \in R^{N^z}$. The code $I \in R^N$ is in the form of a one hot vector in which the desired identity is given by $y^{idx}$ which will be 1 in the one hot vector. The goal of $G$ is to generate realistic looking fake image $\bar{x}$ which can fool $D$ to classify it as fake, the identity $I$ and expression $y^e$ with the following objective function: 
\vspace{0.9mm}
\begin{align}
\underset{G}{\mathrm{max}}{V_G}(D,G) ={}&\underset{z\sim p_z(z),I\sim p_I(I)}{\mathrm{E_{x,y\sim p_e(x,y)}}}[\log({D_{y^e}^e}{(G(x,I,z))}+\notag\\
&\log({D_{y^{idx}}^{id}}{(G(x,I,z))}]\notag\\
\end{align}
The ultimate goal of generator, which strives to transfer expression from input image to output image with a target identity, is to learn a discriminative expression representation $f_{exp} (x)$ which is disentangled from the identity information. The disentanglement process is performed by inputting an identity code $I$ to $G_{de}$, and thus $G_{en}$ is trained to learn only the expression information from the input images because noise $z$ models the other variations like gender, age illumination etc.   

\subsubsection{Discriminator: }
The architecture of our discriminator is different from conventional GAN in such away that we have employed a multi-task CNN. The task of our discriminator is to classify between real and fake images and in addition to that its other task is to classify the identity and expression using the following objective function:
\vspace{0.9mm}
\begin{align}
\underset{G}{\mathrm{max}}{V_G}(D,G)={}&{E_{x,y\sim p_e(x,y)}}[\log({D_{y^e}^e}(x)+\log({D_{y^{id}}^{id}}(x)]+\notag\\
&\underset{z\sim p_z(z),I\sim p_I(I)}{\mathrm{E_{x,y\sim p_e(x,y)}}}[\log({D_{N^e+1}^e}{(G(x,I,z))}]\notag\\
\end{align}

Where $D^{id}$ and $D^e$ corresponds to the identity classification task and expression classification task of our multi-task discriminator respectively. $N^d$ denotes the total number of subjects in the dataset and $N^e$ denotes the number of expressions which in our case is six for MMI and Olulu Casia datasets and seven in case of CK+.  Given a real expression image $x$, the first part of objective function of $D$ is to classify its identity and expression. The second part of the above equation shows that the objective of $D$ is also to maximize the probability of a synthetic image $\bar{x} = G_{de} (f_{id} (x),I,z)$ generated by the generator, being classified as a fake class. The expression classification in the discriminator $D$ helps in transferring expressions from input image to the synthesized generated images. 

\section{Facial Expression Recognition}
After the training of DE-GAN, the disentangled expression representation $f_{exp} (x)$ from input expression image is extracted by using only the encoder of DE-GAN$'$s generator. The classification of facial expression is performed using the extracted expression representation by training a simple shallow classifier like MLP or SVM. In our technique we do not have to extract features from multiple layers of encoder and decoder and train multiple CNNs such as in \cite{r5} to perform FER. We instead use a 350 long one dimensional vector extracted from the last layer of our encoder and train a very shallow multi-layer perceptron for FER. 

\section{Experiments}
The proposed DE-GAN based FER technique is evaluated on three publicly available facial expression databases: I.e CK+ \cite{r20}, Oulu-CASIA \cite{r21} and MMI \cite{r22} database. 

\subsection{Implementation Details}
Facial landmarks are detected by employing Convolutional Experts Constrained Local Model (CE-CLM) \cite{r23}, and face detection and face alignment is performed based on those detected facial landmarks. After face alignment, $75 \times 75$ regions are  randomly sampled from the aligned faces. Data augmentation is applied to avoid the over-fitting problem by increasing the number of training images. In the data augmentation process, five patches of size $75 \times 75$ are cropped-out from five different locations I.e center and four corners of each image. Ten angles i.e $-150^\circ$, $-120^\circ$, $-90^\circ$, $-60^\circ$, $-30^\circ$, $30^\circ$, $60^\circ$, $90^\circ$, $120^\circ$, $150^\circ$ are then used to rotate each image patch. In order to further increase the number of training data each rotated image is horizontally flipped. Thus as a result of this process the original dataset is augmented 110 times. The data augmentation process is not applied to testing data. 

For the optimization of the hyper-parameters the optimization strategies presented in \cite{r24} are adopted in our technique. Adam optimizer \cite{r25} is used with a batch size of 150, learning rate of 0.0001 and momentum of 0.5. Normal distribution is used with a zero mean and standard deviation of 0.02 to initialize all network weights. Contrary to conventional GAN training strategies mentioned in \cite{r3}, in DE-GAN, in later iterations when $D$ reaches to near optimal solution, $G$ is updated more frequently than $D$, due to supervised classification provided by the class labels. 

\subsection{Experimental Results}
\subsubsection{CK+ dataset: }
 Is a famous facial expression recognition database which contains 327 videos sequences from 118 subjects. Each of these sequences corresponds to one of seven expressions, i.e. anger, contempt, disgust, fear, happiness, sadness, and surprise, where each sequence starts from neutral expression to peak expression. The entire CK+ data-base for the training and testing of the proposed method is compiled in such a way that only the last three frames of each sequence is taken as an expression image, which results in 981 images. The result of our experiments on CK+ database is reported in Table 1.    

\begin{table}[t!]
\centering
\begin{tabular}{|l| c| c|} 
 \hline
 Method&Setting&Accuracy\\
 \hline
 \hline
 CNN(baseline)& Static  & 90.34\\
 \textbf{DE-GAN(Ours)}& Static  & \textbf{97.28}\\
 \hline
\end{tabular}
\caption{CK+: Accuracy for seven expressions classification.}
\label{table:1}
\end{table}

\subsubsection{MMI dataset: }
This dataset contains expression images from 31 subjects in the form of 236 video image sequences. Six basic expressions are used to label each of these sequences. The dataset which is used for the evaluation of the proposed technique consists of 208 sequences from 31 subjects in which each frame is captured in frontal view. The expressions in each sequence starts from neutral, evolves into peak expression in the middle of the sequence and dies down to neutral again in the end. The three middle frames which correspond to the peak expression are selected from each sequence to construct a dataset containing 624 images. The accuracy of facial expression recognition on MMI database is shown in Table. 2. 

 \subsubsection{Oulu-CASIA dataset: }
 This dataset consists of three parts corresponding to images obtained using two different cameras in three different lighting environments. In this experiment, only the data compiled under strong illumination condition using the VIS camera is used for training and testing. In Oulu-CASIA VIS 80 subjects are employed to construct a dataset which contains 480 sequences, where each sequence is labeled as one of the six basic expressions. The expressions in each video sequence starts from neutral and ends at peak expression. The last three frames of each sequence is selected to create a training and testing dataset. 

The accuracy of the proposed method on Oulu-CASIA dataset is shown in Table. 3. The accuracy of FER using the disentangled expression features from DE-GAN is high in case of Oulu-CASIA dataset due to the fact that the expression images for each of the six basic expressions is present for each subject in the Oulu-CASIA database. Due to the complete dataset, we are being able to effectively disentangle expression features from the identity features. While in case of CK+ and MMI datasets not all subjects contain all six/seven expression image sequences. Thus it shows that a better disentangled facial expression information can be obtained when the proposed method is performed on datasets where expression information is complete for all subjects in that dataset.

\begin{table}[t!]
\centering
\begin{tabular}{|l| c| c|} 
 \hline
 Method&Setting&Accuracy\\
 \hline
 \hline
 CNN(baseline)& Static  & 58.46\\
 \textbf{DE-GAN(Ours)}& Static  & \textbf{72.97}\\
 \hline
\end{tabular}
\caption{MMI: Accuracy for six expressions classification.}
\label{table:1}
\end{table}

\begin{table}[t!]
\centering
\begin{tabular}{|l| c| c|} 
 \hline
 Method&Setting&Accuracy\\
 \hline
 \hline
 CNN(baseline)& Static  & 73.14\\
 \textbf{DE-GAN(Ours)}& Static  & \textbf{89.17}\\
 \hline
\end{tabular}
\caption{Oulu-CASIA: Accuracy for six expressions classification.}
\label{table:1}
\end{table}

\section{Conclusions}
In this paper we have presented DE-GAN, which is a facial expression recognition method based on learning by synthesis or reconstruction. The main goal of DE-GAN is to extract and disentangle the expression information from the identity information from a facial expression image. In order to achieve this goal an encoder-decoder structured generator is employed in DE-GAN, in which the disentangled expression representation is learned by transferring the expression from input image to a synthesized image with different identity than that of the identity of the input image. The identity of the synthesized image is fed explicitly to the decoder part of DE-GAN. In order to improve the performance of facial expression image synthesis and disentangled expression representation learning, we have used a multi-task CNN based discriminator, whose job is to not only classify between real and fake images, but it also classifies the identity and expression information.  
Initial experimental results evaluated on publicly available databases using the proposed method show that the disentangled expression features learned using the proposed technique is comparable with the results of the state-of-the-art facial expression recognition techniques.

\bibliographystyle{aaai}
\bibliography{bib.bib}

\begin{thebibliography}{}

\bibitem[\protect\citeauthoryear{Baltru{\v{s}}aitis, Mahmoud, and
  Robinson}{2015}]{r9}
Baltru{\v{s}}aitis, T.; Mahmoud, M.; and Robinson, P.
\newblock 2015.
\newblock Cross-dataset learning and person-specific normalisation for
  automatic action unit detection.
\newblock In {\em 2015 11th IEEE International Conference and Workshops on
  Automatic Face and Gesture Recognition (FG)}, volume~6,  1--6.
\newblock IEEE.

\bibitem[\protect\citeauthoryear{Cai \bgroup et al\mbox.\egroup }{2018}]{r19}
Cai, J.; Meng, Z.; Khan, A.~S.; Li, Z.; O’Reilly, J.; and Tong, Y.
\newblock 2018.
\newblock Island loss for learning discriminative features in facial expression
  recognition.
\newblock In {\em 2018 13th IEEE International Conference on Automatic Face \&
  Gesture Recognition (FG 2018)},  302--309.
\newblock IEEE.

\bibitem[\protect\citeauthoryear{Cai \bgroup et al\mbox.\egroup }{2019}]{r2}
Cai, J.; Meng, Z.; Khan, A.~S.; Li, Z.; O'Reilly, J.; and Tong, Y.
\newblock 2019.
\newblock Identity-free facial expression recognition using conditional
  generative adversarial network.
\newblock {\em arXiv preprint arXiv:1903.08051}.

\bibitem[\protect\citeauthoryear{Chen \bgroup et al\mbox.\egroup }{2013}]{r12}
Chen, J.; Liu, X.; Tu, P.; and Aragones, A.
\newblock 2013.
\newblock Learning person-specific models for facial expression and action unit
  recognition.
\newblock {\em Pattern Recognition Letters} 34(15):1964--1970.

\bibitem[\protect\citeauthoryear{Chu, De~la Torre, and Cohn}{2016}]{r10}
Chu, W.-S.; De~la Torre, F.; and Cohn, J.~F.
\newblock 2016.
\newblock Selective transfer machine for personalized facial expression
  analysis.
\newblock {\em IEEE transactions on pattern analysis and machine intelligence}
  39(3):529--545.

\bibitem[\protect\citeauthoryear{Ding, Zhou, and Chellappa}{2017}]{r32}
Ding, H.; Zhou, S.~K.; and Chellappa, R.
\newblock 2017.
\newblock Facenet2expnet: Regularizing a deep face recognition net for
  expression recognition.
\newblock In {\em 2017 12th IEEE International Conference on Automatic Face \&
  Gesture Recognition (FG 2017)},  118--126.
\newblock IEEE.

\bibitem[\protect\citeauthoryear{Goodfellow \bgroup et al\mbox.\egroup
  }{2014}]{r3}
Goodfellow, I.; Pouget-Abadie, J.; Mirza, M.; Xu, B.; Warde-Farley, D.; Ozair,
  S.; Courville, A.; and Bengio, Y.
\newblock 2014.
\newblock Generative adversarial nets.
\newblock In {\em Advances in neural information processing systems},
  2672--2680.

\bibitem[\protect\citeauthoryear{Jiang, Valstar, and Pantic}{2011}]{r14}
Jiang, B.; Valstar, M.~F.; and Pantic, M.
\newblock 2011.
\newblock Action unit detection using sparse appearance descriptors in
  space-time video volumes.
\newblock In {\em Face and Gesture 2011},  314--321.
\newblock IEEE.

\bibitem[\protect\citeauthoryear{Jung \bgroup et al\mbox.\egroup }{2015}]{r30}
Jung, H.; Lee, S.; Yim, J.; Park, S.; and Kim, J.
\newblock 2015.
\newblock Joint fine-tuning in deep neural networks for facial expression
  recognition.
\newblock In {\em Proceedings of the IEEE international conference on computer
  vision},  2983--2991.

\bibitem[\protect\citeauthoryear{Kim \bgroup et al\mbox.\egroup }{2015}]{r15}
Kim, B.-K.; Lee, H.; Roh, J.; and Lee, S.-Y.
\newblock 2015.
\newblock Hierarchical committee of deep cnns with exponentially-weighted
  decision fusion for static facial expression recognition.
\newblock In {\em Proceedings of the 2015 ACM on International Conference on
  Multimodal Interaction},  427--434.
\newblock ACM.

\bibitem[\protect\citeauthoryear{Kingma and Ba}{2014}]{r25}
Kingma, D.~P., and Ba, J.
\newblock 2014.
\newblock Adam: A method for stochastic optimization.
\newblock {\em arXiv preprint arXiv:1412.6980}.

\bibitem[\protect\citeauthoryear{Li and Deng}{2018}]{r8}
Li, S., and Deng, W.
\newblock 2018.
\newblock Deep facial expression recognition: A survey.
\newblock {\em arXiv preprint arXiv:1804.08348}.

\bibitem[\protect\citeauthoryear{Li, Deng, and Du}{2017}]{r18}
Li, S.; Deng, W.; and Du, J.
\newblock 2017.
\newblock Reliable crowdsourcing and deep locality-preserving learning for
  expression recognition in the wild.
\newblock In {\em Proceedings of the IEEE conference on computer vision and
  pattern recognition},  2852--2861.

\bibitem[\protect\citeauthoryear{Liu \bgroup et al\mbox.\egroup }{2014}]{r28}
Liu, M.; Li, S.; Shan, S.; Wang, R.; and Chen, X.
\newblock 2014.
\newblock Deeply learning deformable facial action parts model for dynamic
  expression analysis.
\newblock In {\em Asian conference on computer vision},  143--157.
\newblock Springer.

\bibitem[\protect\citeauthoryear{Lucey \bgroup et al\mbox.\egroup }{2010}]{r20}
Lucey, P.; Cohn, J.~F.; Kanade, T.; Saragih, J.; Ambadar, Z.; and Matthews, I.
\newblock 2010.
\newblock The extended cohn-kanade dataset (ck+): A complete dataset for action
  unit and emotion-specified expression.
\newblock In {\em 2010 IEEE Computer Society Conference on Computer Vision and
  Pattern Recognition-Workshops},  94--101.
\newblock IEEE.

\bibitem[\protect\citeauthoryear{Martinez \bgroup et al\mbox.\egroup
  }{2017}]{r7}
Martinez, B.; Valstar, M.~F.; Jiang, B.; and Pantic, M.
\newblock 2017.
\newblock Automatic analysis of facial actions: A survey.
\newblock {\em IEEE transactions on affective computing}.

\bibitem[\protect\citeauthoryear{Meng \bgroup et al\mbox.\egroup }{2017}]{r4}
Meng, Z.; Liu, P.; Cai, J.; Han, S.; and Tong, Y.
\newblock 2017.
\newblock Identity-aware convolutional neural network for facial expression
  recognition.
\newblock In {\em 2017 12th IEEE International Conference on Automatic Face \&
  Gesture Recognition (FG 2017)},  558--565.
\newblock IEEE.

\bibitem[\protect\citeauthoryear{Ng \bgroup et al\mbox.\egroup }{2015}]{r17}
Ng, H.-W.; Nguyen, V.~D.; Vonikakis, V.; and Winkler, S.
\newblock 2015.
\newblock Deep learning for emotion recognition on small datasets using
  transfer learning.
\newblock In {\em Proceedings of the 2015 ACM on international conference on
  multimodal interaction},  443--449.
\newblock ACM.

\bibitem[\protect\citeauthoryear{Pantic \bgroup et al\mbox.\egroup
  }{2005}]{r22}
Pantic, M.; Valstar, M.; Rademaker, R.; and Maat, L.
\newblock 2005.
\newblock Web-based database for facial expression analysis.
\newblock In {\em 2005 IEEE international conference on multimedia and Expo},
  5--pp.
\newblock IEEE.

\bibitem[\protect\citeauthoryear{Radford, Metz, and Chintala}{2015}]{r24}
Radford, A.; Metz, L.; and Chintala, S.
\newblock 2015.
\newblock Unsupervised representation learning with deep convolutional
  generative adversarial networks.
\newblock {\em arXiv preprint arXiv:1511.06434}.

\bibitem[\protect\citeauthoryear{Sariyanidi, Gunes, and Cavallaro}{2014}]{r6}
Sariyanidi, E.; Gunes, H.; and Cavallaro, A.
\newblock 2014.
\newblock Automatic analysis of facial affect: A survey of registration,
  representation, and recognition.
\newblock {\em IEEE transactions on pattern analysis and machine intelligence}
  37(6):1113--1133.

\bibitem[\protect\citeauthoryear{Tran, Yin, and Liu}{2017}]{r34}
Tran, L.; Yin, X.; and Liu, X.
\newblock 2017.
\newblock Disentangled representation learning gan for pose-invariant face
  recognition.
\newblock In {\em Proceedings of the IEEE Conference on Computer Vision and
  Pattern Recognition},  1415--1424.

\bibitem[\protect\citeauthoryear{Valstar \bgroup et al\mbox.\egroup
  }{2012}]{r13}
Valstar, M.~F.; Mehu, M.; Jiang, B.; Pantic, M.; and Scherer, K.
\newblock 2012.
\newblock Meta-analysis of the first facial expression recognition challenge.
\newblock {\em IEEE Transactions on Systems, Man, and Cybernetics, Part B
  (Cybernetics)} 42(4):966--979.

\bibitem[\protect\citeauthoryear{Yang, Ciftci, and Yin}{2018}]{r5}
Yang, H.; Ciftci, U.; and Yin, L.
\newblock 2018.
\newblock Facial expression recognition by de-expression residue learning.
\newblock In {\em Proceedings of the IEEE Conference on Computer Vision and
  Pattern Recognition},  2168--2177.

\bibitem[\protect\citeauthoryear{Yang, Zhang, and Yin}{2018}]{r1}
Yang, H.; Zhang, Z.; and Yin, L.
\newblock 2018.
\newblock Identity-adaptive facial expression recognition through expression
  regeneration using conditional generative adversarial networks.
\newblock In {\em 2018 13th IEEE International Conference on Automatic Face \&
  Gesture Recognition (FG 2018)},  294--301.
\newblock IEEE.

\bibitem[\protect\citeauthoryear{Yu and Zhang}{2015}]{r16}
Yu, Z., and Zhang, C.
\newblock 2015.
\newblock Image based static facial expression recognition with multiple deep
  network learning.
\newblock In {\em Proceedings of the 2015 ACM on International Conference on
  Multimodal Interaction},  435--442.
\newblock ACM.

\bibitem[\protect\citeauthoryear{Y{\"u}ce, Gao, and Thiran}{2015}]{r11}
Y{\"u}ce, A.; Gao, H.; and Thiran, J.-P.
\newblock 2015.
\newblock Discriminant multi-label manifold embedding for facial action unit
  detection.
\newblock In {\em 2015 11th IEEE International Conference and Workshops on
  Automatic Face and Gesture Recognition (FG)}, volume~6,  1--6.
\newblock IEEE.

\bibitem[\protect\citeauthoryear{Zadeh \bgroup et al\mbox.\egroup }{2017}]{r23}
Zadeh, A.; Chong~Lim, Y.; Baltrusaitis, T.; and Morency, L.-P.
\newblock 2017.
\newblock Convolutional experts constrained local model for 3d facial landmark
  detection.
\newblock In {\em Proceedings of the IEEE International Conference on Computer
  Vision},  2519--2528.

\bibitem[\protect\citeauthoryear{Zhao \bgroup et al\mbox.\egroup }{2011}]{r21}
Zhao, G.; Huang, X.; Taini, M.; Li, S.~Z.; and Pietik{\"a}Inen, M.
\newblock 2011.
\newblock Facial expression recognition from near-infrared videos.
\newblock {\em Image and Vision Computing} 29(9):607--619.

\bibitem[\protect\citeauthoryear{Zhao \bgroup et al\mbox.\egroup }{2016}]{r33}
Zhao, X.; Liang, X.; Liu, L.; Li, T.; Han, Y.; Vasconcelos, N.; and Yan, S.
\newblock 2016.
\newblock Peak-piloted deep network for facial expression recognition.
\newblock In {\em European conference on computer vision},  425--442.
\newblock Springer.

\end{thebibliography}
\end{document}